\documentclass[10pt,twocolumn,letterpaper]{article}

\usepackage[pagenumbers]{iccv} %

\usepackage{multirow}
\usepackage{graphicx}
\usepackage{subcaption}
\usepackage{mwe} %
\usepackage{adjustbox} %
\usepackage{pifont}
\usepackage{calc}            %
\usepackage{varwidth}        %
\usepackage{bbm}

\newcommand{\cmark}{\ding{51}}%

\makeatletter
\newcommand\footnoteref[1]{\protected@xdef\@thefnmark{\ref{#1}}\@footnotemark}
\makeatother

\definecolor{iccvblue}{rgb}{0.21,0.49,0.74}
\usepackage[pagebackref,breaklinks,colorlinks,allcolors=iccvblue]{hyperref}

\def\paperID{7} %
\def\confName{ICCV}
\def\confYear{2025}

\title{From Global to Local: Social Bias Transfer in CLIP}

\author{
Ryan Ramos$^{1}$ \quad Yusuke Hirota$^{2*}$ \quad Yuta Nakashima$^1$ \quad Noa Garcia$^1$ \\
$^1$The University of Osaka \quad $^2$NVIDIA
}

\begin{document}
\maketitle

\def\thefootnote{*}\footnotetext{Work done at The University of Osaka.
}
\renewcommand{\thefootnote}{\arabic{footnote}}

\begin{abstract}
        The recycling of contrastive language-image pre-trained (CLIP) models as backbones for a large number of downstream tasks calls for a thorough analysis of their transferability implications, especially their well-documented reproduction of social biases and human stereotypes.
        How
        do such biases, learned during pre-training, propagate to downstream applications like visual question answering or image captioning?  Do they transfer at all?
        
        We investigate this phenomenon, referred to as bias transfer in prior literature, through a comprehensive empirical analysis. Firstly, we examine how pre-training bias varies between global and local views of data, finding that bias measurement is highly dependent on the subset of data on which it is computed. Secondly, we analyze correlations between biases in the pre-trained models and the downstream tasks across varying levels of pre-training bias, finding difficulty in discovering consistent trends in bias transfer. Finally, we explore why this inconsistency occurs, showing that under the current paradigm, representation spaces of different pre-trained CLIPs tend to converge when adapted for downstream tasks. We hope this work offers valuable insights into bias behavior and informs future research to promote better bias mitigation practices.
    
\end{abstract}

\begin{figure}[t!]
  \centering
  \includegraphics[width=\linewidth]{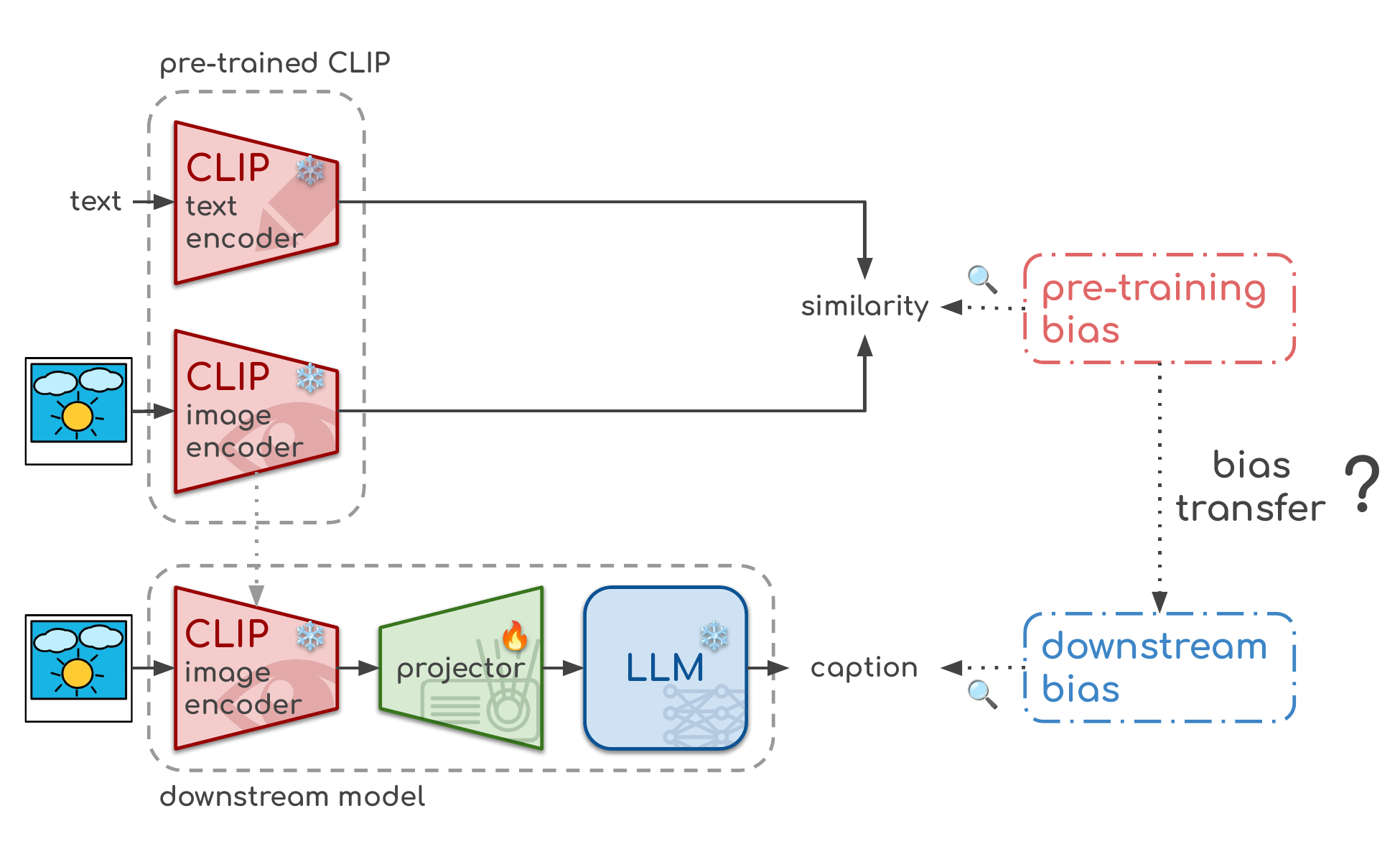}
  \caption{An overview of our method for bias transfer analysis. We first measure pre-training bias in CLIPs. We then use them to train models for downstream tasks and measure the resulting downstream bias. By repeating this with different CLIPs, bias metrics, and protected attributes, we can calculate Spearman rank correlation coefficients to determine how bias transfer may be observed, or if it occurs at all.}
  \label{fig:teaser}
\end{figure}

\section{Introduction}
\label{sec:intro}

CLIP \cite{radford2021learning} models play important roles in vision-capable conversational agents \cite{liu2024visual, young2024yi, ye2023mplug}, image generation \cite{ramesh2021zero, rombach2022high}, real image editing \cite{mokady2022null, brooks2023instructpix2pix}, and other vision-language tasks.
These models,
which are
used
as backbones for downstream tasks, eliminate the need to train from scratch to capture complex vision-language relationships. However, the CLIP representation space contains numerous documented instances of social bias \cite{wang2021gender, wolfe2022markedness}. For example, CLIP has been found to associate images of Asian individuals with alienating words \cite{wolfe2022american}, images of women with traditionally female occupations \cite{mandal2023gender}, and images of young people with criminality \cite{agarwal2021evaluating}.

Given the prevalent use of pre-trained CLIPs as backbones for downstream tasks, the extent to which social biases are embedded in the CLIP representation space calls for a focus on \textit{bias transfer} \cite{steed2022upstream}. Bias transfer refers to how biases in an original pre-trained model affect biases observed in downstream tasks. 
While prior research has investigated bias transfer in text \cite{steed2022upstream}, vision \cite{wang2023overwriting}, and vision-language \cite{cabello2023evaluating, ghate2025biases}, no study has comprehensively examined the dynamics of bias transfer in the context of pre-trained CLIPs and their use as backbones in  downstream tasks. %

In line with this, we explore both pre-training bias in CLIP and its relationship with bias in downstream tasks, such as visual question answering and image captioning. First, we consider that approaches \cite{geyik2019fairness, agarwal2021evaluating} for evaluating CLIP pre-training bias may have further room for exploration. %
While a majority of metrics do stratify their analysis per protected attribute (\eg, gender, age, and ethnicity), these results tend to be reduced to single values to give \textit{global} understandings of bias over an image dataset.
We hypothesize that in embedding spaces, phenomena exist in localities that are not reflected in evaluations over the entire dataset \eg skintone disparity might not persist at identical intensities between a global view and a local view of an image dataset. In order to probe these localities in embedding spaces, we identify and evaluate bias on groups in these spaces.

Secondly, we investigate how
bias in a pre-trained CLIP
may relate to the 
bias observed on
downstream tasks. An overview of our method is shown in \cref{fig:teaser}.
We first train multiple models for downstream tasks by leveraging a collection of pre-trained CLIPs as backbones.
We then measure the different biases of the pre-trained CLIPs and the biases observed on downsteam tasks, and search for correlations. We conduct these experiments across i) multiple methods for measuring social bias in pre-trained models, ii) multiple downstream tasks and corresponding bias measurement methods, and iii) multiple protected attributes.

Lastly, we explore a potential explanation of our observed results based on the current paradigm of adapting pre-trained CLIPs to frozen language models. We support this by comparing inter-model similarity, in terms of embedding spaces, before and after downstream training.

Our findings are summarized as follow:

\begin{itemize}
    \item A single model may exhibit different levels of bias on different groups of embedded datapoints, despite them coming from the same distribution. A model that is less biased than another model on a global view of data may actually be more biased when evaluating a specific locality of the distribution. Our results indicate that social bias in CLIP is not a uniform phenomenon and is dependent on data used to measure such bias. 

    \item There is a difficulty in finding consistent correlation patterns between any of the factors considered in this study, \ie there is no single pre-training bias measurement method, downstream task, or protected attribute that consistently reveals strong correlations between pre-training and downstream bias. Furthermore, a pre-trained model performing better on one demographic compared to another does not necessarily mean that relationship will hold in its downstream counterpart.

    \item Under the current paradigm for training new models for downstream tasks in which pre-trained CLIPs are adapted to frozen language models, the new representation spaces of these models after training will converge regardless of the original pre-trained model, mitigating any effect from pre-training bias on downstream bias.
\end{itemize}

\begin{table*}[t]
  \caption{Summary of the metrics we use to evaluate bias in pre-training and downstream tasks.}
  \vspace{-3pt}
  \label{tab:metrics}
  \setlength{\tabcolsep}{10pt}
  \centering
  \footnotesize
  \begin{tabular}{@{}lccll@{}}
    \toprule
    metric & pre-training & downstream  & bias as & task\\
    \midrule
    KL-divergence of R@$k$ & \cmark & -  & demographic parity & image-text retrieval\\
    MaxSkew@$k$ & \cmark & - & spurious correlations & text-image retrieval \\
    KL-divergence of VQA & - & \cmark & demographic parity & visual question answering \\
    KL-divergence of CIDEr & - & \cmark & demographic parity & image captioning  \\
    DBA & - & \cmark & spurious correlations & visual question answering  \\
    LIC & - & \cmark & spurious correlations & image captioning  \\

  \bottomrule
  \end{tabular}
\end{table*}

\section{Related work}

\paragraph{Social bias in CLIP}
Prior work has shown that CLIP exhibits gender bias \cite{wang2021gender, wolfe2022markedness, mandal2023gender, agarwal2021evaluating}, racial bias \cite{wolfe2022markedness,mandal2023gender,agarwal2021evaluating}, and age bias \cite{wolfe2022markedness,agarwal2021evaluating}. Birhane \etal \cite{birhane2021multimodal} demonstrated that CLIP tends to favor gender-stereotypical image-text pairs. For example, when evaluating images of female astronauts, CLIP matches them more closely with captions of ``a smiling housewife'' than ``an astronaut''. Similarly, Hazirbas \etal \cite{hazirbas2024bias} found that darker-skinned individuals are more likely to be associated with non-human classes. Methods to evaluate social bias can be categorized into two types: \textit{harmful stereotype association} and  \textit{demographic parity}. Harmful stereotype association \cite{agarwal2021evaluating,wolfe2022american,wolfe2023contrastive} examines associations between demographic groups (\eg women) and predefined concepts (\eg occupations) in CLIP's image-text matching. A representative method is MaxSkew@$k$ \cite{geyik2019fairness}, which assesses whether CLIP associates certain demographics more strongly with specific descriptions (\eg \textit{a smart person}). On the other hand, demographic parity \cite{dehdashtian2024utility, sagawa2019distributionally,garcia2023uncurated, hardt2016equality} measures performance disparity across demographic groups, primarily evaluated by text-to-image retrieval for images representing each demographic.
Overall, previous work on CLIP's social bias analyzes the entire distribution of data for probing bias. In contrast, we compare bias from both \textit{global} (\ie the entire distribution of probing data) and \textit{local} (\ie subsets of probing data) perspectives. Compared to previous works \cite{dehdashtian2024fairerclip, gandelsman2023interpreting} that show bias changing between perspectives due to over-/under-representations of demographics with better performance, we explore how performance itself on these demographics changes between views.

\vspace{-12pt}
\paragraph{Bias transfer}
While transfer learning allows a model that has been pre-trained on one task to leverage its learned representations of inputs to perform well on other tasks, studies have raised concerns whether the biases of these models will also transfer in a phenomenon referred to as \textit{bias transfer} \cite{steed2022upstream}.
The rationale for bias transfer research is justified by the findings of fair representation literature \cite{shen2022fair, feldman2015certifying, mcnamara2017provably, zhao2022inherent, madras2018learning}. Shen \etal \cite{shen2022fair} prove that, theoretically, fairer representations lead to fairer downstream task outcomes, and McNamara \etal \cite{mcnamara2017provably} show that this relationship should be predictable. 
However, empirical results vary. Cabello \etal \cite{cabello2023evaluating} demonstrate a lack of impact from pre-training bias on downstream bias. Wang \etal \cite{wang2023overwriting} show that pre-training bias can affect the downstream bias of vision models, but only if fine-tuning data is scarce or biased. Steed \etal \cite{steed2022upstream} similarly observe that it is fine-tuning data bias instead of pre-training bias that strongly affects the downstream bias of text models, however in their experiments, intervening on the fine-tuning dataset bias only affects non-pre-trained models. Ghate \etal \cite{ghate2025biases} show that CLIP biases affect bias in zero-shot tasks, but do not study whether this holds when the CLIP embedding space is altered.
Building on these efforts, we conduct a comprehensive analysis of bias transfer in CLIP, considering multiple protected attributes (\eg, gender, race), bias definitions, metrics, and downstream tasks.

\section{Preliminaries}
We first introduce the definitions and metrics used in our analysis, which are summarized in \cref{tab:metrics}.

\subsection{Definitions}

We list definitions and annotations as follows:

\vspace{-10pt}
\paragraph{bias} we consider two complementary definitions of bias:

\begin{itemize}[noitemsep,topsep=0pt]
    \item \textbf{bias as demographic disparity} the level of dissimilarity
    between performance results across different demographics, \eg a model that performs better on male images than female images has more gender bias than one that performs on both demographics equally well.
    \item \textbf{bias as spurious correlations } the level to which a model spuriously correlates demographics with certain outputs or behaviors.
    For example, Meister \etal \cite{meister2023gender} observe that gender in visual datasets is spuriously correlated with mean image color or pose of the photo subject. %
\end{itemize}

\vspace{-10pt}
\paragraph{pre-training bias} bias observed by analyzing a pre-trained model prior to any alterations to its representation space or weights.

\vspace{-10pt}
\paragraph{downstream bias} bias observed on downstream tasks from models that modify the representation space of a pre-trained model for the purpose of performing these tasks.

\vspace{-10pt}
\paragraph{bias transfer} the phenomenon where biases of a pre-trained model induce biases in its downstream counterpart.

\vspace{-10pt}
\paragraph{protected attribute}
as defined in \cite{barocas-hardt-narayanan}: ``socially salient categories that have served as the basis for unjustified and
systematically adverse treatment in the past."
We refer to the possible values for a given protected attribute as \textit{protected attribute values}, or simply \textit{demographics}.
We define $\mathcal{A} = \{a\}$ as the set of demographics $a$ such that all $a \in \mathcal{A}$ are values of the same protected attribute. For example, when measuring binary skintone bias, $\mathcal{A} = \{\text{lighter}, \text{darker}\}$.

\subsection{Pre-training bias metrics}

We use two main metrics for measuring the original CLIP's level of social bias: %

\vspace{-7pt}
\paragraph{KL-divergence of R@$k$} To measure bias as demographic disparity, we investigate a model's ability to perform text retrieval equally well across different demographics. For each $a \in \mathcal{A}$, we first calculate the rate at which the model retrieves an image's correct caption within the top $k$ retrievals as recall@$k$ (R@$k$). Then, to measure the disparity among the R@k of different $a$, we calculate the KL-divergence between the L1-normalized observed distribution $p_a = r_a / \sum_{a' \in \mathcal{A}} r_{a'}$ of R@$k$'s, where $r_a$ is R@$k$ for $a$, and the ideal equal-performance distribution $q_a = 1/|\mathcal{A}|$, given by:
\begin{equation}
    d_{\text{R}@k}(\mathcal{A}) = \text{KL}(p, q) = \sum_{a \in \mathcal{A}} p_a \log\frac{p_a}{q_a},
    \label{eq:kldiv_of_recall}
\end{equation}
where $p = \{p_a \mid a \in \mathcal{A}\}$ and $q = \{q_a \mid a \in \mathcal{A}\}$.

\vspace{-7pt}
\paragraph{MaxSkew@$k$ \cite{geyik2019fairness}}
To measure bias as spurious correlations, we investigate how much a model associates the images of specific demographics with certain text descriptions. Given a collection of face images and a text description devoid of words related to protected attributes, we take the top-$k$ images retrieved by a model and measure the largest log-normalized ratio between a demographic's proportion of representation among the $k$ images compared to the ideal proportion, formalized as:
\begin{equation}
    \text{MaxSkew@}k = \max_{a \in \mathcal{A}} \log \frac{f_a}{f'_a}
\end{equation}
where 
$f_a$ is the observed proportion of the top-$k$ retrieved images that belong to $a$; and $f'_{a}$ is the ideal proportion, given by the proportion of all images used in the experiment that belong to $a$. In using this metric to quantify a model's level of bias, we calculate the average MaxSkew@$k$ of a model over multiple text descriptions.

\subsection{Downstream bias metrics}

To measure downstream bias,
we seek a diverse representation of downstream vision-language tasks:
1) visual question answering (VQA) \cite{antol2015vqa}; and 2) image captioning \cite{vinyals2015show, karpathy2015deep, fang2015captions} 
, with bias metrics:

\vspace{-10pt}
\paragraph{KL-divergences of standard performance metrics}
To measure bias as demographic disparity in both VQA and image captioning, we measure the disparity in a model's performance between different demographics $a \in \mathcal{A}$. We calculate this disparity through \cref{eq:kldiv_of_recall}. Images are first divided per demographic $a \in \mathcal{A}$. For VQA, we first measure accuracy per demographic, whereas for image captioning we measure CIDEr \cite{vedantam2015cider}. We calculate the disparity among the results via

\begin{equation}
    d_{M}(\mathcal{A}) = \text{KL}(p'_M, q) %
    \label{eq:kldiv_of_metrics}
\end{equation}
where $p'_M = \{p'_{M,a} \mid a \in \mathcal{A}\}$ is a set of L1-normalized scores $p'_{M,a}$ per $a \in \mathcal{A}$ using metric $M$, such as VQA accuracy or CIDEr.

\vspace{-10pt}
\paragraph{Directional bias amplification} We adopt directional bias amplification (DBA) \cite{wang2021directional} following Cabello \etal \cite{cabello2023evaluating} to measure the spurious correlations between a VQA model's answers and the demographic in the image. Removing samples with answers that are binary, numerical, or outside the top 50 most frequent answers, let $\mathcal{D} = \{(x, t, a)\}$ be the dataset of question $x$, ground-truth answer $t$, and demographic $a$, and $\hat{\mathcal{D}} = \{(x, \hat{t}, a)\}$ be the same but with predicted answer $\hat{t}$. We define $\mathcal{T}=\{t\}$ and $\hat{\mathcal{T}} = \{t\}$ as the unique set of answers in $\mathcal{D}$ and $\hat{\mathcal{D}}$ respectively. Let $P_l(a)$ and $P_l(t)$ denote the fraction of questions within $l \in \{\mathcal{D}, \hat{\mathcal{D}}\}$ corresponding to demographic $a \in \mathcal{A}$ or containing answer $t$ respectively.
We calculate a model's DBA, in the direction of protected attribute influencing predictions, as
\begin{equation}
    \text{BiasAmp}_{\mathcal{A}\rightarrow\mathcal{T}} = \frac{1}{|\mathcal{A}| |\mathcal{T}|} \sum_{a, t} (u_{at} \Delta_{at} - (1 - u_{at})\Delta_{at}), 
\end{equation}
where the summation is computed over all combinations of $a$ and $t$, and
\begin{align}
    u_{at} &= \mathbbm{1}[P_\mathcal{D}(a, t) > P_\mathcal{D}(a) P_\text{D}(t)], \\
    \Delta_{at} &= P_{\hat{\mathcal{D}}}(t \mid a) - P_\mathcal{D}(t \mid a),
\end{align}
with $\mathbbm{1}[\cdot]$ being the indicator function. Higher DBA values imply that a model has stronger spurious correlations between its outputs and input image demographics relative to the training data.

\paragraph{LIC} We use LIC \cite{hirota2022quantifying} to measure the spurious correlations between an image captioner's outputs and the demographic of the image. 
Consider datasets $\mathcal{D} = \{(y^*, a)\}$ and $\mathcal{\hat{D}} = \{(\hat{y}, a)\}$ consisting of ground truth and model-generated captions mentioning $a \in \mathcal{A}$, respectively.
Each dataset is used to train separate BERT \cite{devlin2018bert} classifiers $f^*$ and $\hat{f}$ to predict $a$ from a caption. The LIC score for both the ground truth and model-generated captions are calculated as
\begin{equation}
\begin{aligned}
    \text{LIC}_\mathcal{{D}} &= \frac{1}{|\mathcal{D}|} \sum_{(y^*, a) \in \mathcal{D}} s_a^*(y^*)\mathbbm{1}[f^*(y^*)=a] \\
    \text{LIC}_{\hat{\mathcal{D}}} &= \frac{1}{|\hat{\mathcal{D}}|} \sum_{(\hat{y}, a) \in \mathcal{\hat{D}}} \hat{s}_a(\hat{y})\mathbbm{1}[\hat{f}(\hat{y})=a]
\end{aligned}
\end{equation}
where $s_a^*(y^*)$ and $\hat{s}_a^*(y^*)$ are the softmax-normalized logits of $f^*$'s and $\hat{f}$'s predictions for $a$, respectively. For both $\text{LIC}_\mathcal{D}$ and $\text{LIC}_{\hat{\mathcal{D}}}$, the higher the score, the stronger the spurious correlations are between the captions and the protected attribute. To measure how much a model amplifies pre-existing biases in the ground truth, we calculate $\text{LIC} = \text{LIC}_{\hat{\mathcal{D}}} - \text{LIC}_{\mathcal{D}}$, where $LIC > 0$, $LIC < 0$, and $LIC = 0$ imply an increase, decrease, and no change in spurious correlations respectively.

\section{Social bias analysis}

\begin{figure*}[t!]
    \centering
    \begin{subfigure}{0.18\linewidth}
        \centering
        \includegraphics[width=\linewidth]{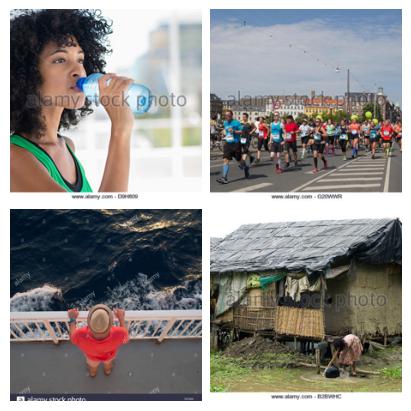}
        \caption{watermarked}
        \label{fig:cluster_a}
    \end{subfigure}
    \begin{subfigure}{0.18\linewidth}
        \centering
        \includegraphics[width=\linewidth]{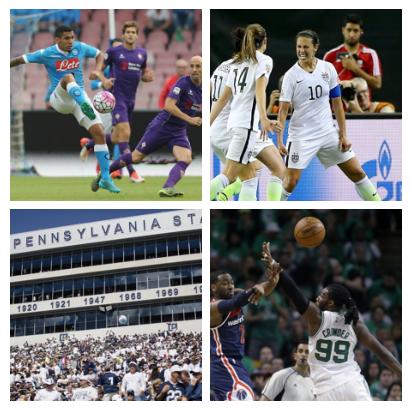}
        \caption{sports}
        \label{fig:cluster_b}
    \end{subfigure}
    \begin{subfigure}{0.18\linewidth}
        \centering
        \includegraphics[width=\linewidth]{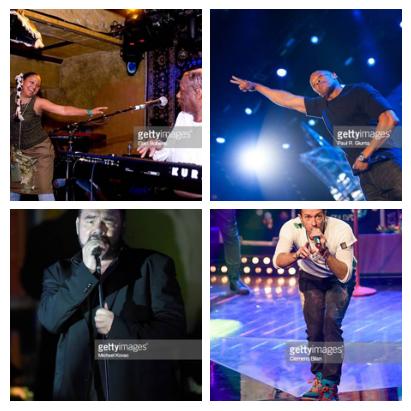}
        \caption{music}
        \label{fig:cluster_c}
    \end{subfigure}
    \begin{subfigure}{0.18\linewidth}
        \centering
        \includegraphics[width=\linewidth]{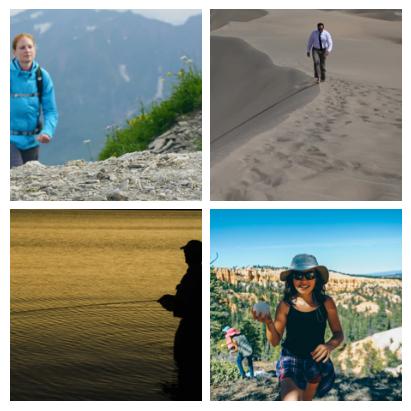}
        \caption{walking}
        \label{fig:cluster_d}
    \end{subfigure}
    \begin{subfigure}{0.18\linewidth}
        \centering
        \includegraphics[width=\linewidth]{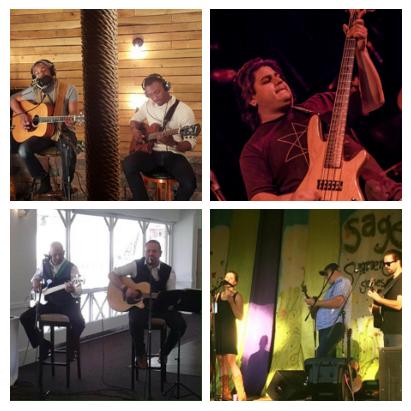}
        \caption{guitars}
        \label{fig:cluster_e}
    \end{subfigure}

    \caption{Images for each group identified in \cref{sec:pt_bias_analysis}, which demonstrate common characteristics that have been identified manually.} 
    \label{fig:groups}
\end{figure*}

\begin{figure*}[t!]
  \centering
  \includegraphics[width=\linewidth]{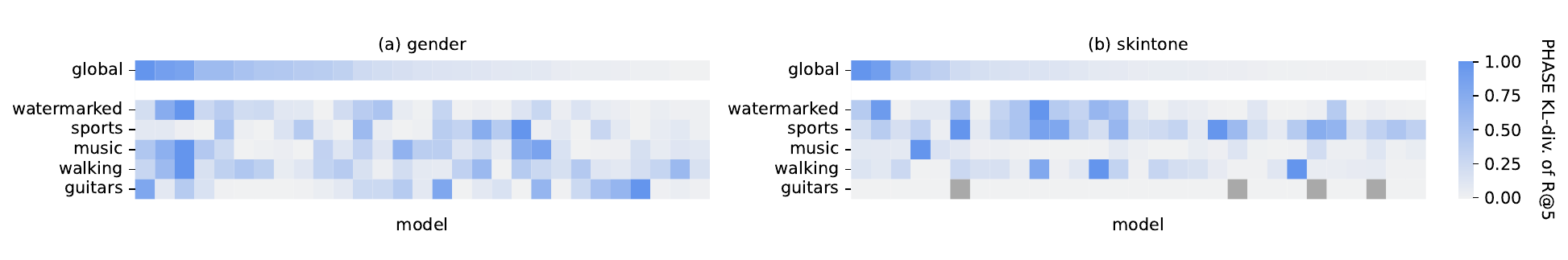}
  \caption{Bias levels observed across models, groups of data, and protected attributes. Darker values indicate stronger bias with lighter values indicating weaker bias. For each protected attribute, the columns are arranged in descending order of bias on the global view (global).
 For visualization purposes, we skip protected attributes dominated by NaN values and normalize values between 0 and 1 per protected attribute.}
  \label{fig:heatmaps}
\end{figure*}

\paragraph{Data} We use five datasets: 1) PHASE \cite{garcia2023uncurated}, 2) COCO \cite{lin2014microsoft}, and 3) FairFace \cite{karkkainen2021fairface} for evaluating pre-training bias, and 4) VQAv2 \cite{goyal2017making}, and 5) COCO captions \cite{chen2015microsoft} for downstream bias. PHASE provides skintone, gender, ethnicity, and age annotations for the $4,614$ image subset of the validation split of GCC \cite{sharma2018conceptual} containing people. The COCO annotations provided by Zhao \etal \cite{zhao2021understanding} cover the $3,756$ image subset of the COCO validation set featuring people, and provides annotations for gender and skintone. Neither PHASE nor the COCO annotations are balanced based on a protected attribute. When using these datasets, we follow prior work \cite{zhao2021understanding,garcia2023uncurated} such that given a certain $\mathcal{A}$, we only use images where all human subjects fall under the same $a \in \mathcal{A}$. FairFace, with the text descriptions from Berg \etal \cite{berg2022prompt} also adopted by prior work \cite{chuang2023debiasing, dehdashtian2024fairerclip}, contains $108,501$ facial images balanced by race, and is accompanied with gender, race, and age annotations. For our downstream metrics, we measure VQA on VQAv2 and image captioning on COCO captions. We combine these with the gender and skintone annotations from Zhao \etal \cite{zhao2021understanding}.

\vspace{-10pt}
\paragraph{Settings}
We measure KL-divergence of R@$k$ on PHASE and COCO and set $k=5$.  MaxSkew@$k$ is measured on FairFace with $k=1000$.

\vspace{-10pt}
\paragraph{Pre-trained models}
\label{sec:models}
We turn to the CLIP model suite provided by Cherti \etal \cite{cherti2023reproducible}, which has 29 CLIPs trained with different configurations based on model architecture, training dataset size, and number of samples seen during training. As shown by Steed \etal \cite{steed2022upstream}, CLIPs trained with different configurations carry naturally different levels of bias.

\subsection{Pre-training analysis: global and local bias}
\label{sec:pt_bias_analysis}

We first investigate CLIP pre-training bias, specifically how bias levels relate between global and local views of the data distribution used for bias probing. 

\begin{table}[t!]
  \caption{Spearman rank correlation coefficients $\rho$, with $p$-values, between local biases and global bias. For illustration purposes we include the correlation coefficients between global biases and themselves, trivially 1, and ignore protected attributes dominated by NaN values.}
  \vspace{-3pt}
  \label{tab:spearman}
  \setlength{\tabcolsep}{4.5pt}
  \centering
  \resizebox{\linewidth}{!}{
  \begin{tabular}{@{}l cc cc cc cc cc cc@{}}
    \toprule
    \multirow{2}{*}{protected attribute} & \multicolumn{2}{c}{global} & \multicolumn{2}{c}{watermarked} & \multicolumn{2}{c}{sports} & \multicolumn{2}{c}{music} & \multicolumn{2}{c}{walking}& \multicolumn{2}{c}{guitars} \\
    \cmidrule(lr){2-3} \cmidrule(lr){4-5} \cmidrule(lr){6-7} \cmidrule(lr){8-9} \cmidrule(lr){10-11} \cmidrule(lr){12-13}
    & $\rho$ & $p$ & $\rho$ & $p$ & $\rho$ & $p$ & $\rho$ & $p$ & $\rho$ & $p$ & $\rho$ & $p$ \\
    \midrule
    gender & 1 & 0.00 &0.59 & 0.00 &-0.01 & 0.95 &0.29 & 0.12 &0.16 & 0.41 &-0.12 & 0.54 \\
    skintone & 1 & 0.00 &0.56 & 0.00 &-0.08 & 0.67 &0.15 & 0.45 &0.23 & 0.23 &NaN & NaN \\
  \bottomrule
  \end{tabular}
  }
\end{table}

\vspace{-7pt}
\paragraph{From global to local space} To study local bias, we divide the entire embedding space into groups. To identify groups, we perform k-means clustering in each embedding space of the 29 pre-trained models, \ie we conduct 29 different k-means clusterings. Then, we match clusters across embedding spaces to define the groups to which an image belongs to. We do this by finding the combination of clusters across different clusterings that provide the maximal cardinality over the intersection of their images. We focus on combinations whose intersection, the resulting groups, contain at least 100 images. We perform this on PHASE with $k=6$ clustering, which results in five groups with at least 100 images each. Each group carries clear and consistent semantic meanings, as shown in \cref{fig:groups}.

\vspace{-14pt}
\paragraph{Results}

\cref{fig:heatmaps} visualizes an example of our results.
Models are sorted from left to right in descending levels of bias observed in the overall dataset, with color intensity indicating stronger bias.
We observe that bias is not a strictly uniformly global phenomenon over data. For example, for gender bias, the model that is the fourth least biased over the entire dataset is the most biased on images containing guitars.
A numerical quantification of this is shown in \cref{tab:spearman}, which shows the Spearman rank correlation coefficients between global and local biases.
With the exception of watermarked photos which have fair to strong\footnote{\label{note:cor_coef}{\url{https://ncbi.nlm.nih.gov/pmc/articles/PMC6107969/table/tbl1/}}} Spearman rank correlation strengths with $p\text{-values}<0.05$, there exists little consistency regarding which groups or protected attributes exhibit similar behaviors. Firstly, no other correlation analysis is statistically significant. Secondly, even if they were, no protected attribute or group other than watermarked photos consistently dominates in terms of having local biases that are more strongly correlated with global bias.
These results reveal that a model's bias is not stable across its embedding space, and subpopulation shifts on a dataset used for probing bias can affect bias measurement.

\begin{figure*}[t!]
    \centering
    \includegraphics[width=0.9\linewidth]{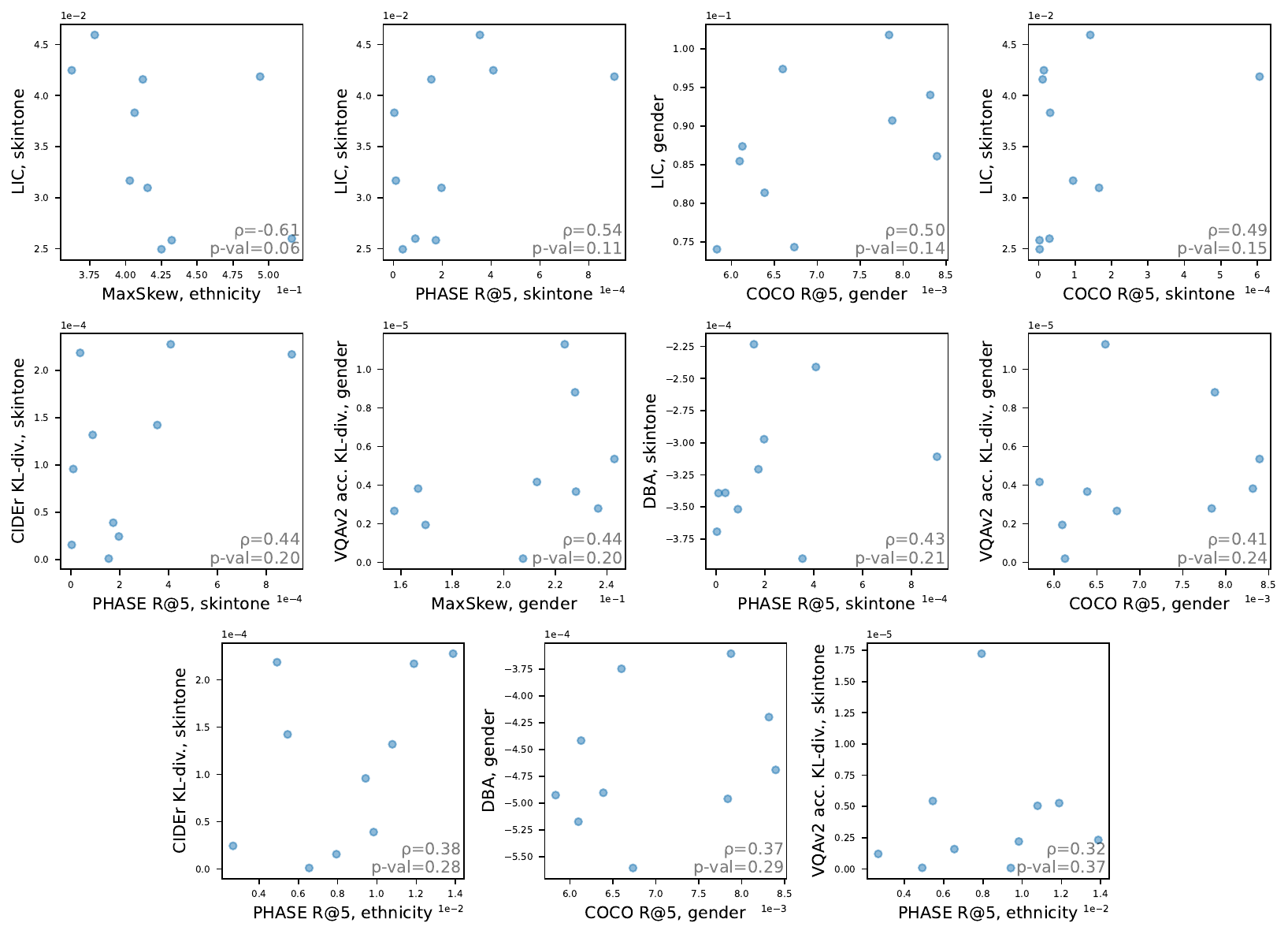}
    \caption{A visualization of our analysis on bias transfer. We show results with Spearman's $\rho \ge 0.3$, the threshold where results can begin to be considered fair or moderate
    .
    The upper left plot shows our strongest result. No other analysis revealed a stronger correlation coefficient or a more significant result.}
    \label{fig:solo_spearman_plot}
\end{figure*}

\begin{figure}[t!]
    \centering
    \includegraphics[width=0.75\linewidth]{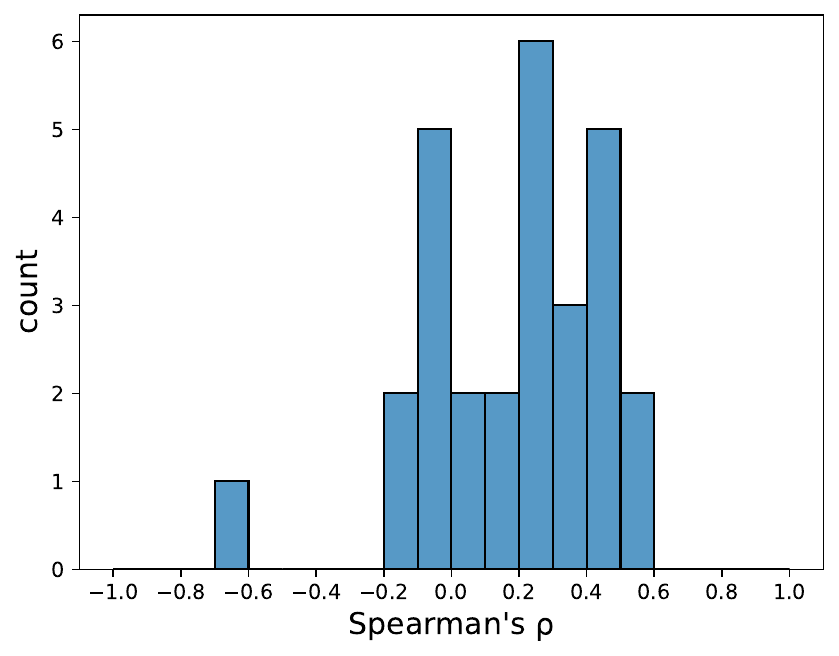}
    \caption{The distribution of our experiments' Spearman's $\rho$ values. Even if our results had $p$-values $< 0.05$, only one experiment reveals strong correlations, with the majority of our correlation strengths considered weak
    .}
    \label{fig:spearman_rho_histogram}
\end{figure}

\begin{figure}[t!]
  \centering
  \includegraphics[width=0.55\linewidth]{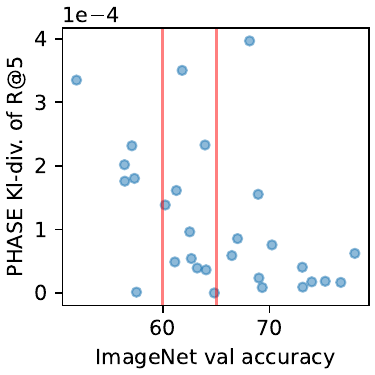}
  \caption{Models selected for downstream analysis, where each model is represented by its general performance and pre-training bias. Models between the red bars are our selections. We seek models with similar general performances but diverse bias levels.}
  \label{fig:subset_selection}
\end{figure}

\subsection{Combinatoric analysis of pre-training bias and downstream bias}
\label{sec:correlation_analysis}
In analyzing the relations between pre-training bias and downstream bias, we calculate the Spearman rank correlation coefficient between every combination of pre-training bias metric, downstream bias metric, and protected attribute, resulting in a total of 28 correlation studies.

\vspace{-10pt}
\paragraph{Training models for downstream tasks}
We first train models for downstream tasks using the pre-trained models as backbones. To isolate the impact of pre-training bias on downstream bias, we minimize the influence of models' general performance.
We take models whose ImageNet val accuracies lie within the $60\%$-$65\%$ range, resulting in 10 models.
This selection process is visualized in \cref{fig:subset_selection}.

We train the models for downstream tasks
by adopting the Tiny-LLaVA method \cite{zhou2024tinyllava}. We follow its base recipe, freezing both a CLIP backbone and a Phi-3 language model \cite{abdin2024phi} and training a two-layer multi-layer perceptron with GELU activation \cite{hendrycks2016gaussian} to project CLIP embeddings to the language model space. Training involves: 1) an image-captioning task using a 558k subset of LAION \cite{schuhmann2022laion}, GCC \cite{sharma2018conceptual}, and SBU \cite{ordonez2011im2text}; and 2) supervised instruction tuning with VQA datasets \cite{hudson2019gqa, singh2019towards, goyal2017making, marino2019ok, schwenk2022okvqa} and image datasets \cite{lin2014microsoft, antol2015vqa} plus synthetic instruction-response pairs. As our sole interest is the effect of pre-training bias on downstream bias, all training data and training hyperparameters are kept consistent between different models, with pre-training bias being the primary variable.

\vspace{-10pt}
\paragraph{Results}

\cref{fig:solo_spearman_plot} shows examples of our bias transfer analysis results. Each point indicates a model, represented by its results on a pre-training bias metric and a downstream bias metric for specific protected attributes. While ethnicity is not necessarily interchangeable with skintone, we include analyses examining the relation between pre-training ethnicity bias and downstream skintone bias, as shown in the upper left plot of \cref{fig:solo_spearman_plot}. This upper left plot represents our strongest result in terms of both correlation strength and statistical significance, which are $0.61$ and $0.06$ respectively. Consequently, none of our correlations results pass the standard 0.05 threshold for statistical significance. Furthermore, none of our analyses produced results that reached correlation coefficients considered very strong.\footnoteref{note:cor_coef} The distribution of our Spearman's $\rho$ values can be seen in \cref{fig:spearman_rho_histogram}, which shows that the majority of our results could only be considered weak at best.
Overall, no pre-training bias metric or downstream bias metric maintains a consistently strong rank correlation when compared to either another specific metric or when used to investigate the bias associated with a specific protected attribute.
This difficulty in finding strong correlations between pre-training bias and downstream bias, despite the size of the combinatoric search space, corroborates previous studies which also noted difficulty in finding correlations between measures of pre-training-bias and downstream bias \cite{steed2022upstream, cabello2023evaluating}.

\subsection{Bias transfer analysis: better pre-training performance does not imply better downstream performance}

\begin{figure}[t!]

  \centering
  \begin{subfigure}[t]{0.48\linewidth}
  \centering
    \includegraphics[width=\linewidth]{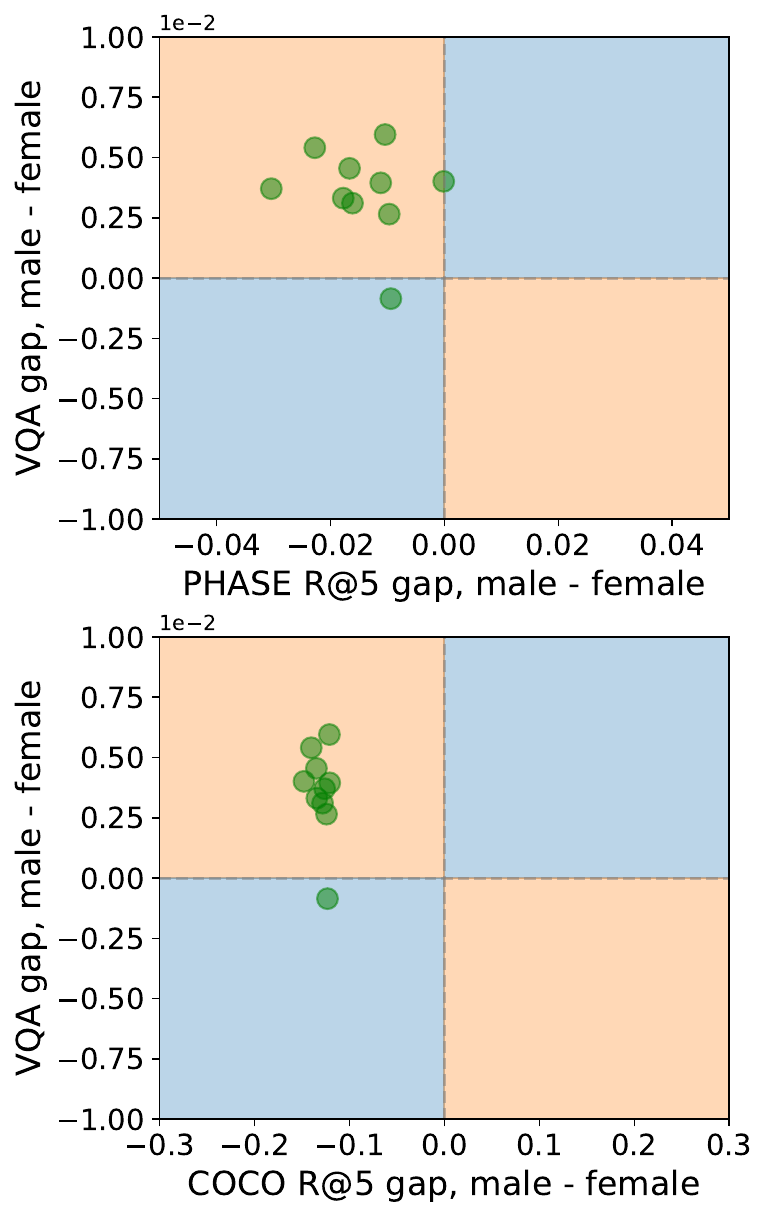}
    \subcaption{R@5 gap vs VQAv2 acc. gap for gender}
    \label{fig:gender_recall_dba_gap}
    \end{subfigure}
    \hfill
  \begin{subfigure}[t]{0.48\linewidth}
  \centering
    \includegraphics[width=\linewidth]{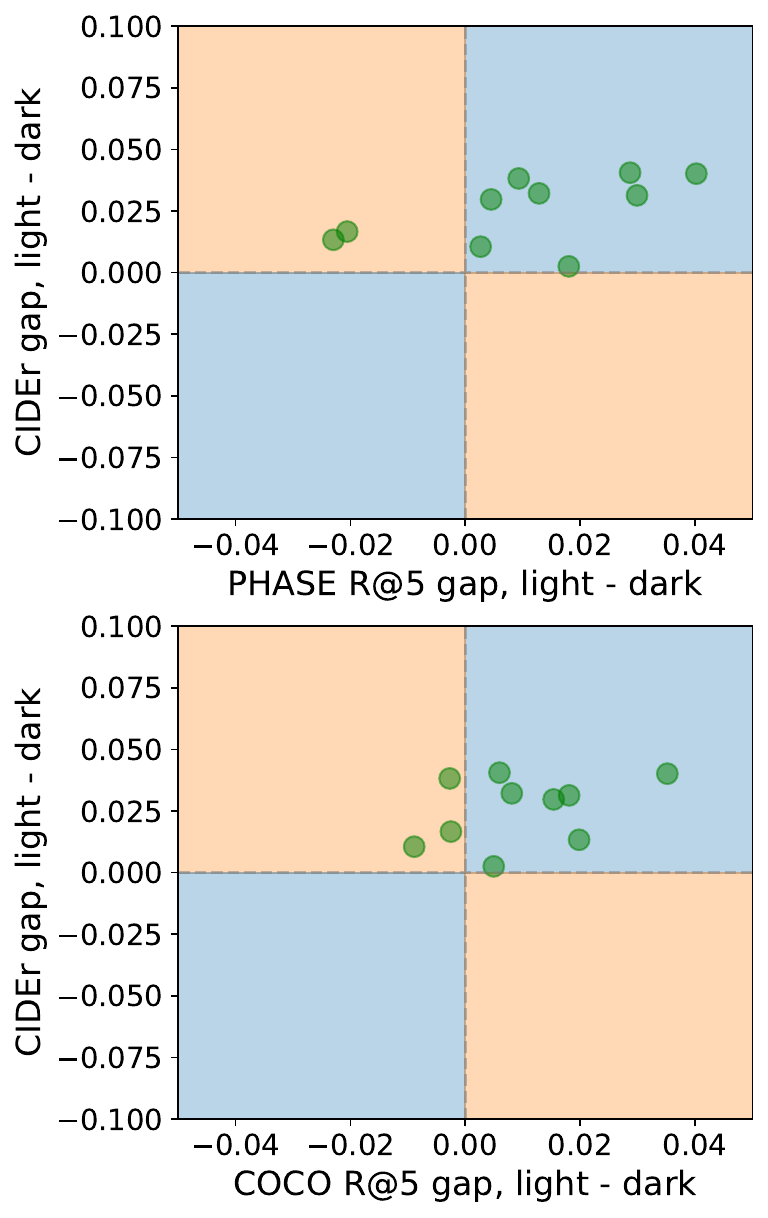}
    \subcaption{R@5 gap vs CIDEr gap for skintone}
    \label{fig:skin_recall_dba_gap}
  \end{subfigure}
  \caption{Comparison between model performances on males and females, and on lighter- and darker-skinned people. X-axes show differences in R@5 values and y-axes show differences in VQAv2 accuracy or CIDEr.}
  \label{fig:recall_dba_gap}
\end{figure}

\begin{figure*}[t]
    \centering
    \begin{subfigure}[c]{0.5\linewidth}
        \centering
        \includegraphics[width=\linewidth]{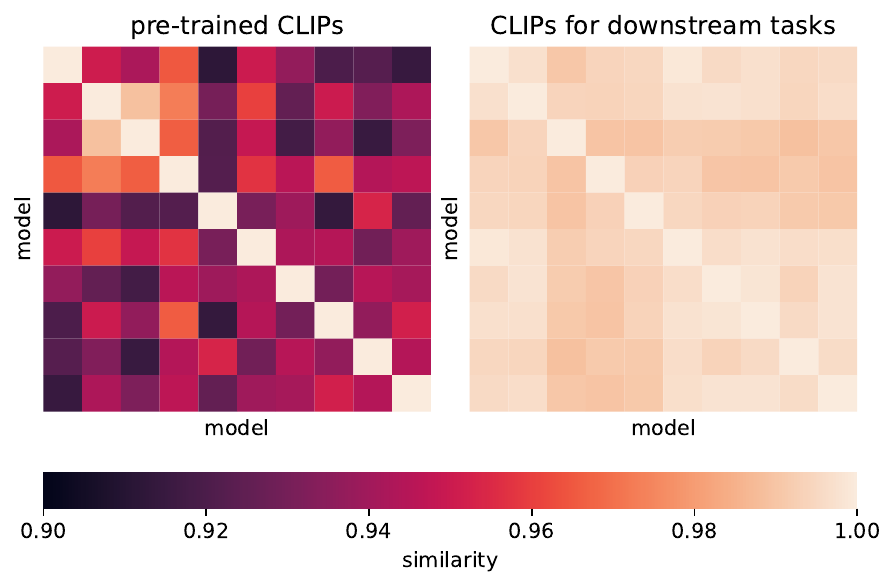}
    \end{subfigure}
    \begin{subfigure}[c]{0.33\linewidth}
        \centering
        \includegraphics[width=\linewidth]{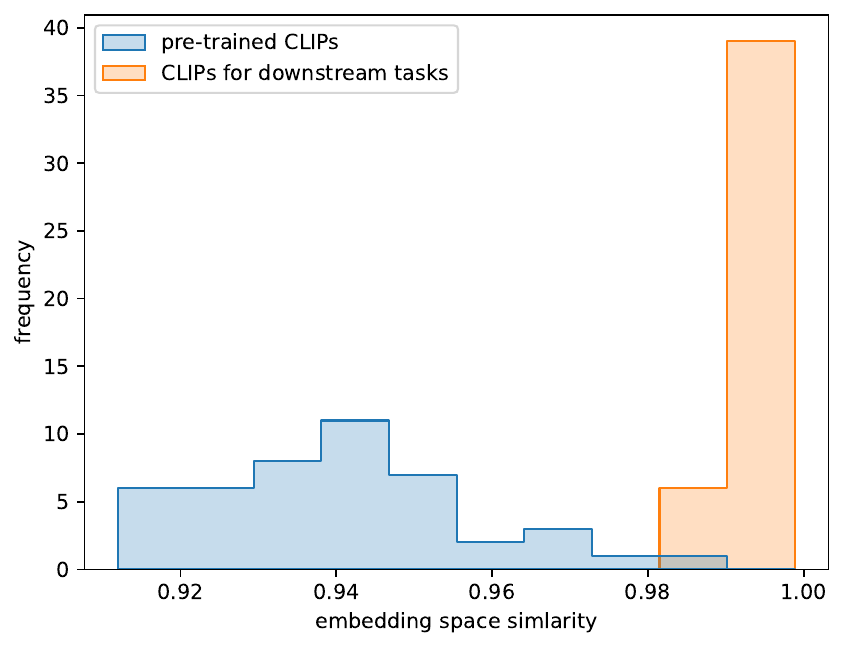}
    \end{subfigure}
    \begin{subfigure}[c]{0.5\linewidth}
        \centering
        \subcaption{Cosine similarity map}
        \label{fig:platonic_heatmaps}
    \end{subfigure}
    \begin{subfigure}[c]{0.3\linewidth}
        \centering
        \subcaption{Histogram of pairwise similarities}
        \label{fig:platonic_histograms}
    \end{subfigure}
    \caption{Similarities between embedding spaces of models, before and after downstream training.}
    \label{fig:both}
\end{figure*}

For a deeper look into how a pre-trained model's performance on a demographic affects its downstream bias, we examine protected attributes that can be divided into two demographics, namely gender and skintone, and compare the two.
We compare the difference in performance by a pre-trained model between the two demographics with the difference in downstream performance of the same two demographics, to investigate whether a demographic being subject to better performance by a pre-trained model will also be subject to better performance downstream.

\vspace{-12pt}
\paragraph{Results}
\cref{fig:recall_dba_gap} shows the relationship between the pre-training performance gap between two demographics and the corresponding downstream gap. If a pre-trained model performing better on one demographic leads to the demographic having better downstream performance, the scatterplot would strictly occupy quadrants I and III, colored in blue. The inverse would imply quadrants II and IV, colored in orange.  We observe a lack of a generally monotonically increasing or decreasing relationship when comparing relative performances between images of different demographics for both pre-training and downstream tasks. A pre-trained model performing better on lighter-skinned images compared to darker-skinned images does not necessarily entail that lighter-skinned images will have a higher VQA accuracy compared to darker-skinned images. There is also a lack of an inverse relation. We observe similar results on gender and captioning.

\subsection{Explanation analysis: frozen components produce similar models and similar biases}
We hypothesize that bias transfer fails to be exhibited consistently due to common paradigms in training multimodal language models, specifically the use of frozen language models \cite{liu2024visual, alayrac2022flamingo}. Because pre-trained models are allowed to adapt to the embedding spaces of language models which are frozen, then regardless of the pre-training model used, sufficient training will eventually cause them to converge to match the language model's embedding space. Consequently, similar models would have similar biases. We argue that what we observe in the downstream bias metrics in \cref{sec:correlation_analysis} is the naturally occurring variance between values that have converged.

To prove this, we compare how similar models' embedding spaces are among the pre-trained models and the models trained for downstream tasks by
computing all pairwise similarities between the $D$ images of a given dataset. With a vector of the $\frac{D (D - 1)}{2}$ similarities produced per model, we
calculate the similarities between
the vectors of similarities of each model. With $E$ as the number of models, this produces $\frac{E (E - 1)}{2}$ similarities. We can compare the $E$ models' inter-similarity before and after downstream training.

We choose all pre-trained models selected in \cref{sec:correlation_analysis}, and their corresponding models for downstream tasks, alongside PHASE for our analysis, with cosine similarity as our similarity metric.
To probe the embedding spaces of pre-trained models after adaption for downstream tasks, we use the trained MLP connectors described in \cref{sec:correlation_analysis} after applying the pre-trained models' respective pooling methods over their penultimate representations, following the original Tiny-LLaVA method.

\vspace{-10pt}
\paragraph{Results}
\cref{fig:platonic_heatmaps} shows similarity maps between models. Note that all cells along the diagonal show the similarity of a model with itself, which is trivially 1. We observe that inter-model similarities increase across all pairs after downstream training. \cref{fig:platonic_histograms} shows the distribution of these similarities, which shifts higher and reduces in variance after downstream training.
The intensity of the shift can be shown quantitatively: we calculate the mean and standard deviation of the pre-trained models and their counterparts for downstream tasks as $\mu_\text{P}=0.940, \sigma_\text{P}=0.017$ and $\mu_\text{D}=0.994, \sigma_\text{D}=0.003$ respectively.
The worst similarity between models for downstream tasks is already $2.89 \times \sigma_\text{P}$ above $\mu_\text{P}$. Furthermore, $\sigma_\text{D}$ is more than $5\times$ smaller than $\sigma_\text{P}$.
The consistent increase in inter-model similarities and reduction in variance supports our theory of model convergence. This corroborates the findings of Wang \etal \cite{wang2023overwriting}, who proved that with a sufficient amount of fine-tuning data, vision models will converge to similar bias levels.

\section{Conclusions}

We investigated CLIP bias across both pre-training and downstream bias. Our analysis of pre-training bias showed that bias is not observed in a uniform manner across CLIP's embedding space. Different localities in an embedding space are subject to different bias levels, indicating that bias evaluation is dependent on the data used for probing. On the connection between pre-training bias and downstream bias, we investigated bias transfer across numerous CLIPs, thus extending bias transfer research towards modern models and fine-tuning paradigms. Our analyses also showed a difficulty in determining rank correlations between pre-training bias and downstream bias, corroborating previous studies \cite{steed2022upstream, cabello2023evaluating,ghate2025biases}.
We further showed that a comparatively better performance on a demographic by a pre-trained model does not always lead to comparatively better performance on downstream tasks for the same demographic.
We explain these results by proving that different pre-trained models will eventually converge to the same representation space when adapted to the same language model, making pre-training bias largely irrelevant to downstream bias.

\section*{Broader Impact}
While the goal of making algorithms fairer is often desirable, debiasing methods can be used by malicious actors to target marginalized individuals in surveillance applications. We do not condone these applications.

\section*{Acknowledgments}
This work was supported by JSPS KAKENHI No. JP23H00497 and JP22K12091, JST CREST Grant No. JPMJCR20D3, and JST FOREST Grant No. JPMJFR216O.

{
    \small
    \bibliographystyle{ieeenat_fullname}
    \bibliography{main}
}

\end{document}